\documentclass[conference]{IEEEtran}

\makeatletter

\def\ps@IEEEtitlepagestyle{%
  \def\@oddfoot{}%
  \def\@evenfoot{}%
}

\usepackage{blindtext}
\usepackage{eso-pic}
\usepackage{cite}
\usepackage{amsmath,amssymb,amsfonts}
\usepackage{algorithmic}
\usepackage{graphicx}
\usepackage{textcomp}
\usepackage{xcolor}
\def\BibTeX{{\rm B\kern-.05em{\sc i\kern-.025em b}\kern-.08em
    T\kern-.1667em\lower.7ex\hbox{E}\kern-.125emX}}
    
\usepackage{eso-pic}

\begin{document}
\title{\vspace*{1cm} A Comparison of Deep Learning Architectures for Optical Galaxy Morphology Classification\\
}

\author{\IEEEauthorblockN{Ezra Fielding}
\IEEEauthorblockA{\textit{Department of Computer Science} \\
\textit{University of the Western Cape}\\
Cape Town, South Africa \\
0000-0002-7936-0222}
\and
\IEEEauthorblockN{Clement N. Nyirenda}
\IEEEauthorblockA{\textit{Department of Computer Science} \\
\textit{University of the Western Cape}\\
Cape Town, South Africa \\
0000-0002-4181-0478}
\and
\IEEEauthorblockN{Mattia Vaccari}
\IEEEauthorblockA{\textit{Department of Physics \& Astronomy} \\
\textit{University of the Western Cape}\\
Cape Town, South Africa \\
0000-0002-6748-0577}
}

\maketitle
\begin{abstract}
The classification of galaxy morphology plays a crucial role in understanding galaxy formation and evolution. Traditionally, this process is done manually. The emergence of deep learning techniques has given room for the automation of this process. As such, this paper offers a comparison of deep learning architectures to determine which is best suited for optical galaxy morphology classification. Adapting the model training method proposed by Walmsley \textit{et al} in 2021, the \textit{Zoobot} Python library is used to train models to predict Galaxy Zoo DECaLS decision tree responses, made by volunteers, using EfficientNet B0, DenseNet121 and ResNet50 as core model architectures. The predicted results are then used to generate accuracy metrics per decision tree question to determine architecture performance. DenseNet121 was found to produce the best results, in terms of accuracy, with a reasonable training time. In future, further testing with more deep learning architectures could prove beneficial. 
\end{abstract}


\begin{IEEEkeywords}
Deep Learning, Optical Galaxy Morphology, Classification, Astronomy
\end{IEEEkeywords}

\section{Introduction}
Morphology plays a crucial role in the investigation of galaxy formation and evolution \cite{walmsley2021galaxy}. As such, accurate and detailed morphological classifications are required to advance research in this area. This process of classification has historically been performed visually by scientists. However, modern astronomical surveys have been producing more detailed images than scientists can visually classify \cite{walmsley2021galaxy}. Projects, such as Galaxy Zoo, have attempted to solve this problem by asking so called ‘citizen scientists’, volunteer members of the public, to provide classifications through a web interface \cite{lintott2008galaxy}. While these efforts have been largely successful and have greatly benefited the scientific community, it is only a matter of time until the output of modern astronomical surveys exceeds efforts which rely solely on human intelligence \cite{masters2019twelve, galvin2020cataloguing}. In other words, projects which rely on humans to visually classify morphology, will not be able to keep up with the sheer volume of data produced by next generation surveys, such as the Square Kilometer Array (SKA), resulting in missed discoveries and insights \cite{dewdney2009square,fluke2020surveying}.

Different approaches have been used to solve the problem of classifying galaxies according to their morphology. In the original Galaxy Zoo study, approximately one hundred thousand participants collectively provided more than 4×$10^7$ individual classifications for nearly one million galaxies from the Sloan Digital Sky Survey (SDSS) \cite{lintott2008galaxy}. Volunteers submitted classifications on the Galaxy Zoo website, where they were shown an image cutout of an area of sky centered on a galaxy. These raw classifications were then aggregated and used to form a final catalogue.  These Galaxy Zoo results were consistent  with those for a subset of SDSS classified by professional astronomers \cite{lintott2008galaxy}. Subsequent projects such as Radio Galaxy Zoo and Galaxy Zoo DECaLS have also found success in using ‘citizen scientists’, to visually classify and inspect galaxies \cite{banfield2015radio, walmsley2021galaxy}. The results of these Galaxy Zoo projects have been used as the basis for a large number of studies on galaxy morphology \cite{masters2019twelve}.

However, despite these successes, as the amount of survey data in need of classification continues to increase, approaches relying mainly on human classification and inspection will require too much time and effort to be viable \cite{galvin2020cataloguing}. For example, the recent Galaxy Zoo DECaLS project would have taken approximately 8 years to collect the standard 40 classifications for each of the 311488 suitable galaxies with the current number of volunteers available \cite{walmsley2021galaxy}. By that time, new surveys would have begun. This highlights the need for classification methods which do not rely heavily on human input.

A fair amount of work has been done in implementing deep learning algorithms for the purpose of classifying galaxy morphology \cite{fluke2020surveying}. A large portion of this work makes use of artificial neural networks to predict galaxy morphology based on images. These networks are trained using a labeled subset of existing surveys. Convolutional neural networks (CNNs) have been shown to be particularly useful and accurate in the classification of astronomical objects, provided the sufficient availability of labeled datasets \cite{dieleman2015rotation, alhassan2018first, dominguez2018improving, burke2019deblending, lukic2018radio, ma2019machine}. Of particular interest is the work of Becker \textit{et al}, which comparatively investigates the performance of different CNN architectures applied to radio galaxy classification. In their work, MCRGNET, RADIO GALAXY ZOO, and CONVXPRESS (a novel classifier) were the architectures which had the best balance of computational requirements and recognition performance \cite{becker2021cnn}. Walmsley \textit{et al} used Galaxy Zoo DECaLS response data to train an EfficientNet B0 architecture to predict the responses to the GDZ-5 decision tree used by the Galaxy Zoo project to classify galaxies \cite{walmsley2021galaxy}.

Most existing works have focused on comparing state-of-the-art deep learning architectures on radio-spectrum images. In terms of optical-spectrum images, the work in \cite{walmsley2021galaxy}, which is the leading one, only uses the EfficientNet B0 and does not present a comparison with other prominent deep learning architectures. Therefore, this paper aims to assess the performance of three different, popular deep learning architectures in predicting volunteer vote fractions for optical-spectrum galaxy images from the recent Galaxy Zoo DECaLS project. EfficientNet B0 \cite{tan2019efficientnet}, DenseNet121 \cite{huang2017densely} and ResNet50 \cite{he2016deep} will compared in this work. Performance criteria will include training time and accuracy metrics such as precision, recall and F1-score. 

The rest of this paper is organised as follows. Section II presents a brief overview of the deep learning architectures used in this work. Section III presents the materials and methods used in this work, while Section IV presents the results and their discussion. Section V draws the conclusion.

\section{Deep Learning Architectures Used in this Work}
Three popular deep learning architectures were selected for this comparison. DenseNet121 and ResNet50 were selected alongside EfficientNet B0, which is the default architecture included with Walsmley's \textit{Zoobot} library \cite{zoobot}. These architectures were chosen based on their size, popularity, and ease of integration into the existing \textit{Zoobot} library code base. These architectures are briefly described as follows.

\paragraph{EfficientNet B0}EfficientNet is a CNN-based architecture that uses a scaling method that uniformly scales all dimensions of depth, width, and resolution using a compound coefficient. The EfficientNet family of models are, therefore, optimised for both accuracy and efficiency, with the aim of performing better than traditional CNNs \cite{tan2019efficientnet}. Apart from the classification of galaxies, EfficientNet has been applied to many other areas such as Covid-19 diagnosis \cite{marques2020automated}, automated fruit recognition \cite{duong2020automated} and multi-label classification of fundus images \cite{wang2020multi}. 

\paragraph{DenseNet121}The DenseNet family of models differentiates itself from traditional CNN's by connecting each layer to every other layer in a feed-forward fashion \cite{huang2017densely}. This works to reduce the number of parameters and alleviate the vanishing gradient problem. DenseNets achieve high performance while requiring less memory and computation when compared to other CNNs. DenseNet has also been applied to many areas such as remote sensing image classification \cite{tong2020channel}, malware detection \cite{hemalatha2021efficient}, and classification of Alzheimer's disease \cite{ruiz20203d}.

\paragraph{ResNet50}The ResNet Family of models reformulate the model layers as learning residual functions with reference to the layer inputs. This results in models which are easier to optimise and which show a gain in accuracy to increased model depth \cite{he2016deep}. Some of the recent applications include crime prediction \cite{matereke2021comparative, wang2019deep} and skin cancer classification \cite{gouda2020skin}.

Table~\ref{base} presents the generic model that illustrates how the architectures were implemented as layers in the \textit{Zoobot} library's default model. The top fully connected layer for each architecture was removed, and replaced with by the three augmentation layers which feed directly in to the network.

\begin{table}[htbp]
\caption{Generic Model Architecture}
\begin{center}
\begin{tabular}{|c|c|}
\hline
\textbf{Layer (type)} & \textbf{Output Shape} \\
\hline
RandomRotation&(None, 300, 300, 1)\\
RandomFlip&(None, 300, 300, 1)\\
RandomCrop&(None, 224, 224, 1)\\
Core Architecture$^{\mathrm{a}}$&(None, 7, 7, )\\
GlobalAveragePooling2D&(None, )\\
Dense&(None, 34)\\
\hline
\multicolumn{2}{l}{$^{\mathrm{a}}$This layer will be swapped out to test each architecture.}
\end{tabular}
\label{base}
\end{center}
\end{table}

With the top fully connected layer removed, EfficientNet B0 resulted in approximately 4.0 million trainable parameters, while DenseNet121 and ResNet50 resulted in approximately 7.0 million and 23.6 million trainable parameters respectively. The final dense layer of the core model architecture was modified to give predictions smoothly, from 1 to 100 using softmax activation. This was done in order to provide plausible Dirichlet posteriors for galaxy morphology \cite{walmsley2021galaxy}.

\section{Materials and Methods}
The method used for this architecture comparison, follows closely with that presented by Walmsley \textit{et al}, 2021 for predicting Galaxy Zoo GZD-5 decision tree volunteer responses \cite{walmsley2021galaxy}. However, instead of training three models per architecture to form an ensemble network, only one model per architecture was trained. The training was also conducted at a decreased batch size of 64. A modified version of the \textit{Zoobot} library was used for data pre-processing, model creation, model training, and making predictions \cite{zoobot}.

\subsection{Description of the Dataset}
The Galaxy Zoo DECaLS dataset includes 253286 424x424 pixel color images (3 channels in the g, r, and z bands) \cite{walmsley2021galaxy}. Volunteer response data is provided for each galaxy. This response data includes the total votes and vote fractions for each question in the GZD-5 decision tree. Any galaxy with less than 3 volunteer classifications was removed resulting in a 1.5\% reduction in the dataset size. This left a total of 249581 images for training and testing. A standard random 80/20 train-test split was applied to the dataset. This resulted in a training dataset containing 199664 galaxies and a test dataset containing 49917 galaxies. The test dataset will serve as the validation dataset.

\begin{figure}[htbp]
\centerline{\includegraphics[width=\linewidth]{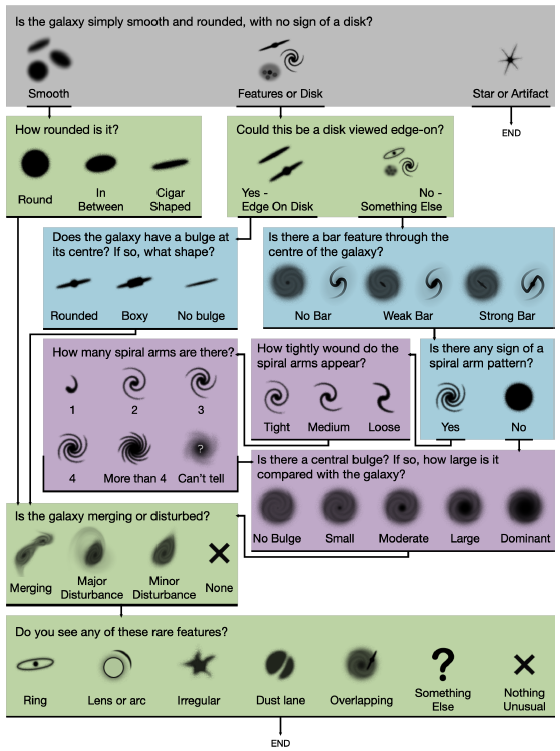}}
\caption{The GZD-5 Decision Tree \cite{walmsley2021galaxy}.}
\label{gzd5}
\end{figure}

The GZD-5 decision tree used for this dataset includes ten questions which have a different number of options depending on the question. A chart explaining the GZD-5 decision tree can be found in Fig.~\ref{gzd5}.

\subsection{Data Processing}
The color information of each image was removed by taking an average over its 3 channels to avoid biasing the morphology predictions \cite{walmsley2020galaxy}. The images were then resized and saved as 300x300x1 matrices. Further augmentations were done when images were loaded into the model. Random horizontal and vertical flips were first applied, followed by a random rotation in the range (0, $\pi$). The image was finally cropped to a size of 224x224 pixels about a random centroid. These augmentations were applied as the images were fed into the model.

\subsection{Training and Testing}
Tensorflow and Python 3.8 were used for model training. The three models were identically trained using the Adam optimizer and loss was calculated by taking the negative log likelihood of observing volunteer answers to all questions from Dirichlet-Multinomial distributions \cite{walmsley2021galaxy}. Training was done with a batch size of 64 and ended after the model loss failed to improve after 10 consecutive epochs. A batch size of 64 was chosen due to the memory limitations of the training hardware. The test dataset was repeated once to reduce variance from random dropouts and augmentations. Training was completed on the South African Center for High Performance Computing's (CHPC) Legau Cluster on two NVIDIA V100 GPUs. Training prototyping and debugging, along with model prediction and evaluation was performed on the ilifu cloud computing facility.

To test and evaluate model performance, a new 80/20 train-test split was applied to the dataset. The new test dataset which resulted from this split was used for evaluation. A galaxy was only used to evaluate a question if 50\% or more of the volunteers shown that galaxy was asked that specific question. Five forward passes per image were made with each model to account for random dropout configurations and augmentations. This resulted in five sets of predicted Dirichlet-Multinomial posteriors per image per model. The mean posterior of the total votes for each model were then calculated and recorded, to be used for calculating the predicted vote fraction for each question option.

The highest predicted vote fraction for a question was rounded to 1, while the remaining fractions were rounded to 0. This was done to produce a set of discrete classifications for each of the GZD-5 decision tree questions. The discrete classifications for each question were then used to generate accuracy metrics, namely precision, recall and F1-score.
\paragraph{Precision}This is defined as the ratio of images that was correctly defined as belonging to the positive class, to the total number of images that were classified as belonging to the positive class.
\paragraph{Recall}This refers to the ratio of the number of images that were correctly classified as belonging to the positive class, to the total number of images in the positive class.
\paragraph{F1-Score}This is the weighted average of recall and precision.

The weighted average metrics are considered for each question. The weighted average metrics are calculated by calculating the metrics for each question option, and finding their average weighted by support. This allows the accuracy metrics to account for label imbalance. Table~\ref{support} shows the support for the metrics of each decision tree question.
\begin{table}[htbp]
\caption{Question Support}
\begin{center}
\begin{tabular}{|c|c|}
\hline
\textbf{Question}&{\textbf{Support}} \\
\hline
smooth-or-featured&49917 \\
disk-edge-on&15445 \\
has-spiral-arms&11380 \\
bar&11380 \\
bulge-size&11380 \\
how-rounded&32526 \\
edge-on-bulge&2475 \\
spiral-winding&7499 \\
spiral-arm-count&7499 \\
merging&49247 \\
\hline
\end{tabular}
\label{support}
\end{center}
\end{table}

\section{Results and Discussion}
The following results were produced after running the experiment. Table~\ref{trainstats} shows the total number of epochs run during training and the total training time.
\begin{table}[htbp]
\caption{Statistics from Training}
\begin{center}
\begin{tabular}{|c|c|c|c|}
\hline
& \textbf{\textit{EfficientNet B0}}& \textbf{\textit{DenseNet121}}& \textbf{\textit{ResNet50}} \\
\hline
\textbf{Total Epochs}&52&37&37 \\
\textbf{Total Time (hours)}&19.795&11.723&9.642 \\
\hline
\end{tabular}
\label{trainstats}
\end{center}
\end{table}

DenseNet121 and ResNet50 both ran for 37 epochs, while EfficientNet B0 ran for 52 epochs. Despite being the largest model by number of parameters, ResNet50 trained for a total of 9.642 hours, while DenseNet121 completed training after 11.723 hours. EfficientNet B0 spent the longest time training, at 19.795 hours.

The weighted average metrics for each question for each of the three models were placed into tables for comparison. Table~\ref{precision} presents a comparison of the precision metric, Table~\ref{recall} presents a comparison of the recall metric, and Table~\ref{f1} presents a comparison of the F1-score metric. The weighted average for all questions is displayed for each model at the bottom of each comparison table.

\begin{table}[htbp]
\caption{Precision Comparison}
\begin{center}
\begin{tabular}{|c|c|c|c|}
\hline
\textbf{Question}&\multicolumn{3}{|c|}{\textbf{Core Model Architecture}} \\
\cline{2-4} 
\textbf{Name} & \textbf{\textit{EfficientNet B0}}& \textbf{\textit{DenseNet121}}& \textbf{\textit{ResNet50}} \\
\hline
smooth-or-featured&0.877&\textbf{0.880}&\underline{0.869} \\
disk-edge-on&\underline{0.954}&\textbf{0.957}&0.955 \\
has-spiral-arms&0.891&\textbf{0.893}&\underline{0.868} \\
bar&0.697&\textbf{0.698}&\underline{0.673} \\
bulge-size&0.684&\textbf{0.690}&\underline{0.675} \\
how-rounded&0.870&\textbf{0.876}&\underline{0.867} \\
edge-on-bulge&0.794&\textbf{0.826}&\underline{0.785} \\
spiral-winding&0.685&\textbf{0.695}&\underline{0.677} \\
spiral-arm-count&0.664&\textbf{0.695}&\underline{0.652} \\
merging&0.849&\textbf{0.851}&\underline{0.845} \\
\hline
\textbf{Weighted Average}&0.838&\textbf{0.843}&\underline{0.831} \\
\hline
\multicolumn{4}{l}{Best model result highlighted in bold. Worst model result underlined.}
\end{tabular}
\label{precision}
\end{center}
\end{table}

\begin{table}[htbp]
\caption{Recall Comparison}
\begin{center}
\begin{tabular}{|c|c|c|c|}
\hline
\textbf{Question}&\multicolumn{3}{|c|}{\textbf{Core Model Architecture}} \\
\cline{2-4} 
\textbf{Name} & \textbf{\textit{EfficientNet B0}}& \textbf{\textit{DenseNet121}}& \textbf{\textit{ResNet50}} \\
\hline
smooth-or-featured&0.876&\textbf{0.881}&\underline{0.863} \\
disk-edge-on&\underline{0.955}&\textbf{0.957}&0.955 \\
has-spiral-arms&0.887&\textbf{0.888}&\underline{0.875} \\
bar&\textbf{0.715}&0.710&\underline{0.696} \\
bulge-size&0.694&\textbf{0.696}&\underline{0.691} \\
how-rounded&0.867&\textbf{0.875}&\underline{0.864} \\
edge-on-bulge&0.820&\textbf{0.825}&\underline{0.816} \\
spiral-winding&0.698&\textbf{0.707}&\underline{0.690} \\
spiral-arm-count&\textbf{0.676}&0.674&\underline{0.664} \\
merging&0.881&\textbf{0.882}&\underline{0.880} \\
\hline
\textbf{Weighted Average}&0.848&\textbf{0.851}&\underline{0.841} \\
\hline
\multicolumn{4}{l}{Best model result highlighted in bold. Worst model result underlined.}
\end{tabular}
\label{recall}
\end{center}
\end{table}

\begin{table}[htbp]
\caption{F1-Score Comparison}
\begin{center}
\begin{tabular}{|c|c|c|c|}
\hline
\textbf{Question}&\multicolumn{3}{|c|}{\textbf{Core Model Architecture}} \\
\cline{2-4} 
\textbf{Name} & \textbf{\textit{EfficientNet B0}}& \textbf{\textit{DenseNet121}}& \textbf{\textit{ResNet50}} \\
\hline
smooth-or-featured&0.876&\textbf{0.881}&\underline{0.865} \\
disk-edge-on&\underline{0.955}&\textbf{0.956}&0.955 \\
has-spiral-arms&0.889&\textbf{0.890}&\underline{0.869} \\
bar&\textbf{0.693}&0.677&\underline{0.662} \\
bulge-size&\underline{0.669}&\textbf{0.674}&0.672 \\
how-rounded&0.867&\textbf{0.875}&\underline{0.865} \\
edge-on-bulge&0.787&\textbf{0.794}&\underline{0.784} \\
spiral-winding&0.687&\textbf{0.696}&\underline{0.678} \\
spiral-arm-count&0.654&\textbf{0.668}&\underline{0.642} \\
merging&0.851&\textbf{0.855}&\underline{0.848} \\
\hline
\textbf{Weighted Average}&0.836&\textbf{0.840}&\underline{0.829} \\
\hline
\multicolumn{4}{l}{Best model result highlighted in bold. Worst model result underlined.}
\end{tabular}
\label{f1}
\end{center}
\end{table}

From the model metric comparison tables, as far as accuracy is concerned, it is clear that DenseNet121 produces the best predictions for the given dataset. ResNet50 generally performs the worst out of the three, and EfficientNet B0 lies in the middle between the two in terms of accuracy.

Taking into account accuracy and training time, DenseNet121 appears to be the most optimal architecture for this dataset. While ResNet50 had the fastest training time, it generally performed the worst for each question in each metric. EfficientNet B0, on the other hand, performed second best in terms of accuracy. However, it took significantly longer to train compared to the other two architectures. DenseNet121 provides the best accuracy performance at a reasonable model training time.

\section{Conclusion}
In conclusion, three popular deep learning architectures (EfficientNet B0, DenseNet121, and ResNet50) were evaluated and compared in terms of accuracy and training performance with regards to classifying optical galaxy morphology.  Performance criteria included training time and accuracy metrics such as precision, recall and F1-score. The method for training outlined by Walmsley \textit{et al} in \cite{walmsley2021galaxy} was followed with some minor changes. In terms of accuracy, DenseNet121 was the best performing architecture with the weighted averages of 0.843, 0.851 and 0.840 for precision, recall and F1-score respectively. On the other hand, ResNet50 exhibited the lowest performance with the weighted average scores of 0.831, 0.841, and 0.829 for precision, recall and F1-score respectively. The training time for DenseNet121 was 11.723 hours, which was approximately 2 hours more than ResNet50. On the other hand, EfficientNet B0 took 19.795 hours. Taking accuracy, training time and total parameters into account, DenseNet121 is the best model to use for this use case.

In future, further testing with a larger selection of deep learning architectures could prove useful and beneficial for the  morphology classification of optical galaxies. Applying these architectures to other Galaxy Zoo projects could also prove insightful.

\section*{Acknowledgment}
The authors would like to thank the Galaxy Zoo volunteers, whose efforts made this dataset possible. Their efforts are individually acknowledged at https://authors.galaxyzoo.org/. Special thanks to Mike Walmsley for granting access to the Galaxy Zoo DECaLS dataset and making the \textit{Zoobot} library publicly available. The authors acknowledge the Centre for High Performance Computing (CHPC), South Africa, and the ilifu cloud computing facility – www.ilifu.ac.za - for providing computational resources to this research project.

\bibliographystyle{IEEEtran}
\bibliography{paperbib}

\end{document}